\documentclass{article}

\PassOptionsToPackage{numbers, compress}{natbib}

\usepackage[preprint]{DICE_arxiv}

\usepackage[utf8]{inputenc} 
\usepackage[T1]{fontenc}    
\usepackage{url}            
\usepackage{booktabs}       
\usepackage{amsfonts}       
\usepackage{amsmath}
\usepackage{nicefrac}       
\usepackage{microtype}      
\usepackage{xcolor}         
\usepackage{algorithm}
\usepackage{multirow}
\usepackage{graphicx}

\usepackage{hyperref}
\usepackage{cleveref}

\title{DICE: Device-level Integrated Circuits Encoder\\with Graph Contrastive Pretraining}

\makeatletter
\renewcommand{\@fnsymbol}[1]{\@arabic{#1}}
\makeatother

\author{
  Sungyoung Lee\thanks{Electrical and Computer Engineering, The University of Texas at Austin} \\
  \And
  Ziyi Wang\thanks{Computer Science and Engineering, The Chinese University of Hong Kong} \\
  \And
  Seunggeun Kim\footnotemark[1] \\
  \And
  Taekyun Lee\footnotemark[1] \\
  \And
  Yao Lai\thanks{Computer Science, The University of Hong Kong} \\
  \And
  David Z. Pan\footnotemark[1]
}

\begin{document}

\maketitle

\begin{abstract}
Pretraining models with unsupervised graph representation learning has led to significant advancements in domains such as social network analysis, molecular design, and electronic design automation (EDA). However, prior work in EDA has mainly focused on pretraining models for digital circuits, overlooking analog and mixed-signal circuits. To bridge this gap, we introduce DICE: Device-level Integrated Circuits Encoder—the first graph neural network (GNN) pretrained via self-supervised learning specifically tailored for graph-level prediction tasks in both analog and digital circuits. DICE adopts a simulation-free pretraining approach based on graph contrastive learning, leveraging two novel graph augmentation techniques. Experimental results demonstrate substantial performance improvements across three downstream tasks, highlighting the effectiveness of DICE for both analog and digital circuits. Our code is available at: \href{https://github.com/brianlsy98/DICE}{https://github.com/brianlsy98/DICE}.
\end{abstract}
\section{Introduction} \label{sec:intro}

Pretraining models through unsupervised learning has demonstrated significant value across various domains, including large language models (LLMs) in language processing~\cite{gpt, devlin2019bert}, vision models in computer vision~\cite{lu2019vilbert, radford2021clip}, and graph models in molecular science~\cite{ross2022large, zeng2022accurate}.
These models are applied to various downstream tasks that differ from their original pretraining objectives and consistently exhibit remarkable performance.
In particular, pretraining with unsupervised graph representation learning has emerged as a powerful methodology in graph-related fields such as social network analysis~\cite{velivckovic2018deep, sun2019infograph}, healthcare~\cite{chowdhury2019mixed, li2022graphbiohealth, seki2024clinical}, molecular science~\cite{liu2019n, wang2022molecular, zhang2024molecular}, and electronic design automation (EDA)~\cite{chen2024dawn, wang2022fgnn, wang2024fgnn2, shi2023deepgate2}.
Such approaches expand the scope and utility of available graph data through integration with domain knowledge, enabling models to learn robust representations independent of specific tasks.

Integrated circuits (ICs) can be effectively represented as graphs, where circuit components (e.g., gates, transistors, or other devices) and their electrical connections (e.g., wires or nets) are modeled as nodes and edges.
This graph-based abstraction enables the application of graph learning methods to EDA in ICs.
However, obtaining labeled data in the IC domain is particularly challenging. This is mainly due to the high costs and time required for running complete IC design and validation processes, the limited number of publicly available circuit designs for dataset construction, and the complexity of the design space, which depends on numerous device parameters.
To enable pretraining without labels, prior work has explored unsupervised graph representation learning methods, where circuits are represented as graphs with logic gates as nodes and their connections as edges.
These approaches have advanced EDA in digital circuit design~\cite{wang2022fgnn, wang2024fgnn2, shi2023deepgate2, shi2024deepgate3, zheng2025deepgate4}.

However, analog circuits cannot be represented using logic gates; instead, they must be described in terms of device components such as transistors, resistors, and capacitors.
Moreover, logic gates themselves are composed of these fundamental devices.
\textbf{Thus, constructing circuit graphs at the device level offers a more foundational approach to pretraining models applicable to both analog and digital ICs. In this work, we address this fundamental yet underexplored problem.}

\begin{figure}[t]
    \centering
    \includegraphics[width=0.95\linewidth]{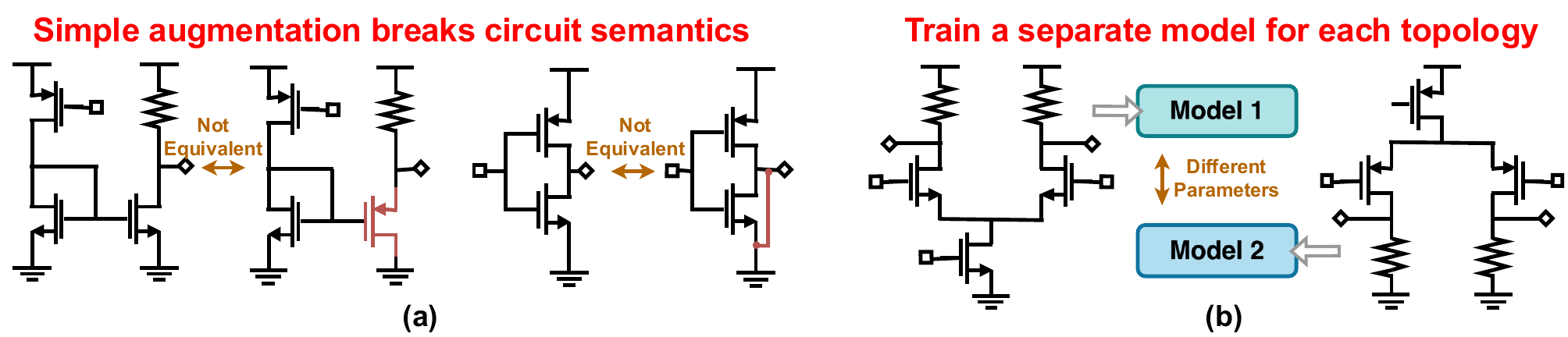}
    \caption{
\textbf{Key challenges for graph contrastive learning on device-level circuits.} (a) Simple graph augmentations, such as randomly adding edges, can break circuit semantics and result in non-equivalent circuits. (b) Existing approaches require training separate models for each circuit topology, limiting the ability to generalize across diverse structures.
}
    \label{fig:problem}
\vskip -0.1in
\end{figure}

To effectively apply unsupervised graph representation learning at the device level, a substantial amount of unlabeled circuit data is required. However, even unlabeled data is scarce, as circuits are typically designed using commercial tools that restrict public availability. To address this, graph contrastive learning~\cite{zhu2021empirical, ju2024towards, graphcl, infogcl, gclaa} offers a promising solution by generating diverse data through augmentation. Nevertheless, two critical challenges arise when applying it to device-level circuits.

\textbf{First, traditional graph augmentation techniques—such as randomly adding nodes or edges—are inadequate for device-level circuits.}
For example, as illustrated in \Cref{fig:problem}(a), simply adding an edge to ground can fix the output voltage at zero, entirely disrupting the original circuit semantics. 
While certain specialized tools~\cite{wolf2013yosys, brayton2010abc} have been effectively used to augment gate-level digital circuits~\cite{wang2022fgnn, li2022deepgate}, these approaches do not apply to device-level circuit augmentation.

\textbf{Second, there is a lack of suitable benchmarks for device-level circuits that incorporate diverse circuit structures within each task.}
As shown in \Cref{fig:problem}(b), prior work typically trains models using a single circuit topology per task~\cite{wang2020gcnrl, settaluri2020autockt, budak2021dnn, poddar2024insight, wang2024learn}, leading to inefficiencies when multiple topologies correspond to the same task. For instance, when predicting identical performance metrics across different operational amplifier circuits, existing approaches require training separate models for each topology.
This limitation in current benchmarks and methodologies hinders the evaluation of pretrained graph models, which should be able to extract robust features across diverse topologies.

Considering these challenges, we propose a graph contrastive learning framework for pretraining \textbf{DICE: Device-level Integrated Circuits Encoder}—a graph neural network (GNN) designed for solving graph-level downstream tasks in both analog and digital ICs.
\Cref{fig:overview} provides an overview of our framework.
We first convert raw circuit files into graph data using our proposed circuit-to-graph mapping method.
Next, we expand the pretraining dataset using two novel data augmentation techniques: positive and negative augmentations, specifically designed for device-level circuits.
Then, DICE is pretrained via contrastive learning on graph-level features.
After pretraining, DICE is integrated into a full message passing neural network (MPNN) and evaluated on our proposed benchmark. Our benchmark includes three downstream tasks, each involving multiple circuit topologies.

The main contributions of our work are as follows:
\begin{itemize}
    \item To the best of our knowledge, we propose the first self-supervised graph representation learning framework applicable to both analog and digital circuits. In contrast to prior work that focuses solely on digital circuits, our approach provides a more fundamental method for pretraining graph models that generalize to both analog and digital ICs.
    \item We introduce two novel data augmentation techniques: positive augmentation, which preserves, and negative augmentation, which perturbs the graph-level functionality of circuits. Both techniques are simulation-free and do not rely on any EDA tools, yet they ensure that the augmented circuits remain valid for simulation by preserving circuit connectivity.
    \item We present a benchmark comprising three graph-level prediction tasks and demonstrate that DICE is effective for both analog and digital circuits. The first task spans both analog and digital domains, the second focuses on digital circuit operations, and the third targets analog circuit operations. None of the tasks is restricted to a single circuit topology, allowing for evaluation across diverse topologies.
\end{itemize}
\section{Preliminaries} \label{sec:prelim}

\subsection{Graph Contrastive Learning}
Graph contrastive learning is one of the major class of unsupervised graph representation learning methods. These approaches learn node, edge, or graph embeddings by maximizing the similarity between positive (similar) samples and minimizing it between negative (dissimilar) samples.
To learn robust representations applicable to various tasks, they rely on data augmentation to diversify training samples.
Augmentations for graph include adding nodes or perturbing edges within the graph.

One of the most widely used objectives in contrastive learning is the NT-Xent loss~\cite{chen2020simple}.
Given a dataset of $N$ samples, each sample is augmented twice to produce $2N$ samples, forming $N$ positive pairs.
For each sample $i \in {1, 2, \ldots, 2N}$, let $\mathbf{z}_i$ denote its embedding and $\mathbf{z}_{j(i)}$ denote the embedding of its positive counterpart.
The NT-Xent loss is then defined as:
\begin{equation}
\small
\label{eq:simclr}
\mathcal{L}_{\text{NT-Xent}}
=
\frac{1}{2N}
\sum_{i=1}^{2N}
-\log
\frac{\exp\bigl(\mathrm{sim}(\mathbf{z}_i, \mathbf{z}_{j(i)}) / \tau\bigr)}
{\sum_{k=1}^{2N} \mathbf{1}_{[k \neq i]}\exp\bigl(\mathrm{sim}(\mathbf{z}_i, \mathbf{z}_k) / \tau\bigr)}
\end{equation}
$\mathrm{sim}(\cdot,\cdot)$ denotes a similarity function (e.g., cosine similarity), and $\tau$ is a temperature coefficient, typically set to 0.1.
Minimizing Eq.~\eqref{eq:simclr} encourages the embeddings of positive pairs to cluster in the feature space while pushing apart the embeddings of negative pairs.

There are also approaches that maximize the similarity between positive pairs without relying on negative samples. SimSiam~\cite{chen2021exploring} is one such method, and the objective is given by:
\begin{equation}
\small
\label{eq:simsiam}
\mathcal{L}_{\text{SimSiam}} = 
-\frac{1}{2} \left(
    \frac{\mathbf{p}_1^\top \, \text{sg}(\mathbf{z}_2)}{\|\mathbf{p}_1\|_2 \cdot \|\text{sg}(\mathbf{z}_2)\|_2}
    +
    \frac{\mathbf{p}_2^\top \, \text{sg}(\mathbf{z}_1)}{\|\mathbf{p}_2\|_2 \cdot \|\text{sg}(\mathbf{z}_1)\|_2}
\right)
\end{equation}
Here, $\mathbf{p}_1$ and $\mathbf{p}_2$ are the outputs of the predictor MLP given the feature inputs $\mathbf{z}_1$ and $\mathbf{z}_2$, respectively.
The operator $\text{sg}(\cdot)$ denotes the stop-gradient operation, and $\|\cdot\|_2$ represents the L2 norm.
The pair $(\mathbf{z}_1, \mathbf{z}_2)$ always constitutes a positive pair, and Eq.~\eqref{eq:simsiam} maximizes their cosine similarity accordingly.

In graph-based settings, contrastive learning can be applied at three levels: node, edge, and graph.
Depending on the target task, the inputs $\mathbf{z}$ in Eq.~\eqref{eq:simclr}-~\eqref{eq:simsiam} can represent node, edge, or graph embeddings.
In this work, we pretrained DICE using graph embeddings.

\subsection{Graph Representation Learning for ICs}

Digital circuits can be fully constructed using logic gates, and they form directed acyclic graphs (DAGs)~\cite{mishchenko2006dag}.
With these graphs, pioneering works have pretrained models via self-supervised learning to reduce the excessive time and cost associated with large-scale digital hardware design.
Methods in \cite{shi2023deepgate2, shi2024deepgate3} pretrain models by comparing logic functionalities and then transferring the models to downstream tasks such as logic synthesis and boolean satisfiability (SAT) solving.
Moreover, graph contrastive learning is employed in \cite{wang2022fgnn, wang2024fgnn2} to pretrain models that cluster functionally equivalent circuits, with applications to downstream tasks such as arithmetic block identification and netlist classification.
To address data scarcity, these works utilize logic synthesis tools such as Yosys~\cite{wolf2013yosys} and ABC~\cite{brayton2010abc}, which generate augmented circuit views that preserve identical functionalities.
Recently, LLMs have also been explored as an approach for augmenting digital circuit data~\cite{chang2024data, liu2023verilogeval}, and these models have been further utilized for multimodal circuit representation learning~\cite{fang2025circuitfusion}.

Analog and mixed-signal (AMS) circuits must be represented using device components, as they cannot be constructed solely with logic gates.
Using device-level circuit graphs, the supervised learning approach in~\cite{hakhamaneshi2022pretraining} pretrains a GNN to predict DC voltage outputs, addressing a node-level prediction task for analog circuits.
The learned representations are subsequently transferred to predict other simulation metrics, such as gain and bandwidth of operational amplifier circuits.
However, self-supervised pretraining methods remain underexplored, in part due to the lack of reliable data augmentation techniques for AMS circuits.
Although prior work has proposed augmentation strategies for AMS circuits~\cite{deeb2023robust, deeb2024graph}, these augmentations are task-specific and primarily focused on analog circuits.
LLMs offer an alternative, as they are capable of generating AMS circuits at the device level~\cite{lai2024analogcoder, vungarala2024spicepilot, bhandari2024auto}, but their effectiveness as an augmentation tool remains uncertain.
\section{Method} \label{sec:method}

Building a graph contrastive learning framework for device-level circuits presents several challenges.
First, it requires a suitable graph construction method that transforms device-level circuits into graph-structured data.
Second, an effective graph augmentation technique that incorporates IC-specific domain knowledge is essential.
Third, a comprehensive contrastive learning framework for pretraining, along with diverse downstream tasks for evaluation, must be established.
We review work related to these challenges in \Cref{appendix:related_works}.

Addressing these challenges, we propose a graph contrastive learning framework for pretraining \textbf{DICE: Device-level Integrated Circuits Encoder.}
In the following subsections, we first introduce our graph construction method for device-level circuits (\Cref{sec:constr}), then describe our customized data augmentation approach that leverages circuit-specific domain knowledge (\Cref{sec:aug}).
Next, we detail our contrastive learning process using the proposed augmentation scheme (\Cref{sec:loss}), and explain how we integrate the pretrained model DICE to solve three downstream tasks (\Cref{sec:dwnstrm}).

\begin{figure}[h!]
    \centering
    \includegraphics[width=\linewidth]{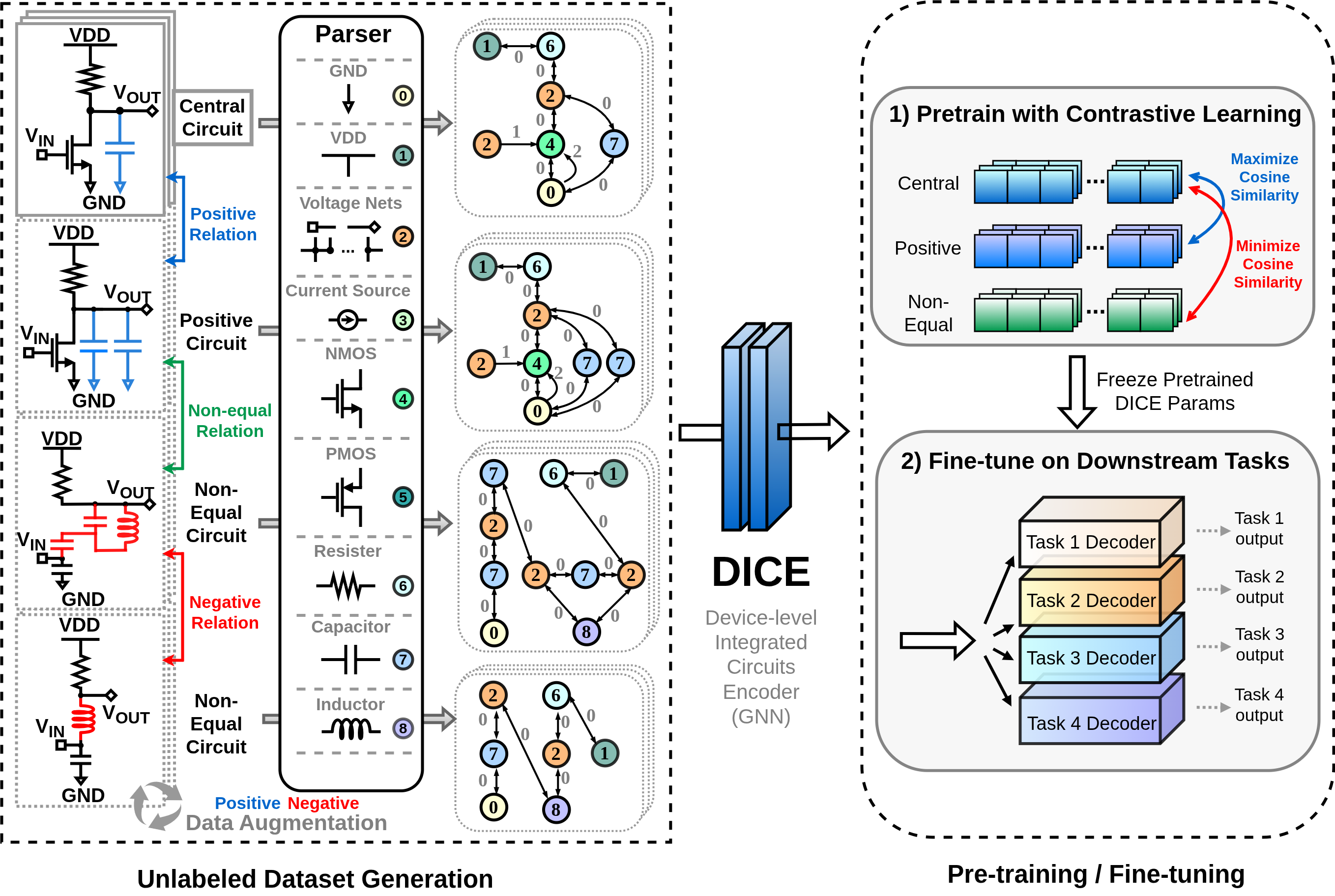}
    \caption{\textbf{Overview of our graph contrastive learning framework for pretraining DICE.} We maximize the cosine similarity between circuits that are in positive relation and minimize it between circuits with a non-equal relation. A detailed explanation of these relations is provided in \Cref{sec:loss}.}
    \label{fig:overview}
\vskip -0.1in
\end{figure}

\subsection{Graph Construction} \label{sec:constr}

Considering prior work stated in \Cref{appendix:motiv_graph_construct}, we propose a new method converting device-level circuit into graph.
The left part of \Cref{fig:overview} illustrates our method.
We begin by using one-hot encoding to represent various device types and their connections, forming a homogeneous graph.
There are nine distinct node types: ground voltage net (0), power voltage net (1), other voltage nets (2), current source (3), NMOS transistor (4), PMOS transistor (5), resistor (6), capacitor (7), and inductor (8).
The relationships among these devices are categorized into five edge types: current flow path (0); connections from voltage nets to NMOS gates (1) or bulks (2); and connections from voltage nets to PMOS gates (3) or bulks (4).
Notably, the edges from gate and bulk voltage nets to MOSFETs are directed, as these nets exert significantly greater influence on the device than vice versa.
With this design, our graph construction method fully leverages edge features without separating device pin nodes from their corresponding device nodes.

\subsection{Positive and Negative Data Augmentation} \label{sec:aug}

\begin{figure*}[t!]
    \begin{center}
    \centerline{\includegraphics[width=\linewidth]{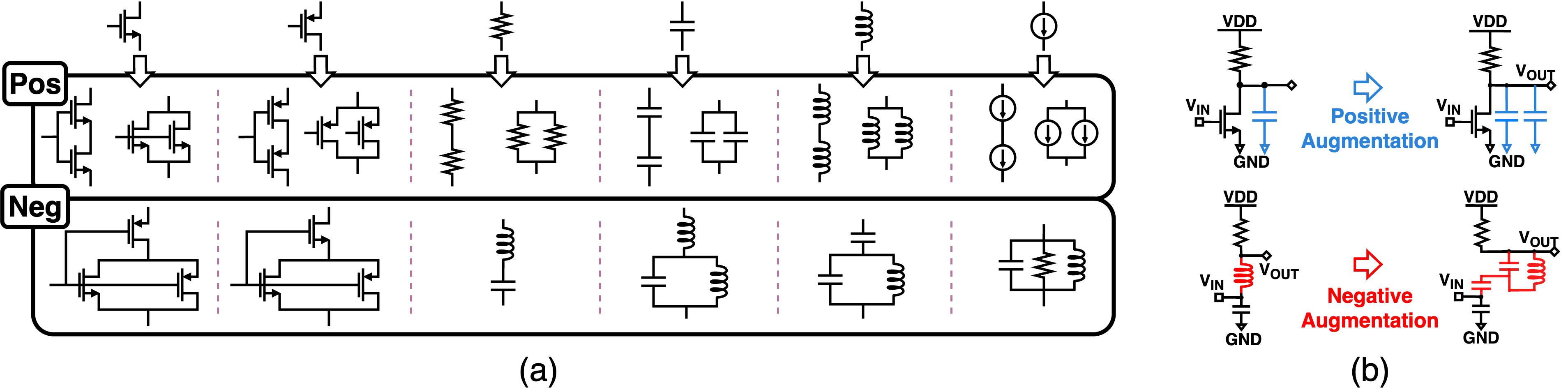}}
    \caption{\textbf{Rules (a) and examples (b) for positive and negative data augmentation.}
    Positive augmentation adds an identical device in parallel or series, preserving the overall circuit function.
    Negative augmentation replaces a subgraph, resulting in a circuit with different overall functionality.
    }
    \label{fig:dataaugmentation}
    \end{center}
    \vskip -0.3in
\end{figure*}

To address the limitations of including device parameters in the pretraining dataset (see \Cref{appendix:motiv_data_aug}), we introduce two data augmentation methods: positive and negative augmentation.
These methods are used to generate the pretraining dataset without incorporating device parameters.
In both methods, subgraphs corresponding to devices within each circuit are modified, as illustrated in \Cref{fig:dataaugmentation}(a).

Positive augmentation generates new circuit topologies while preserving the graph-level functionality of the original circuit.
In each iteration, a random device is selected, and an identical device is added either in parallel or in series.
Although device parameters are not explicitly considered, this modification is functionally equivalent to altering the parameters of the selected device (see \Cref{appendix:posda}).
As a result, the augmented graphs retain graph-level properties that closely resemble those of the original.
Samples generated through positive augmentation can be further augmented to produce additional positive or negative samples.
By maximizing the similarity between these positive samples, DICE learns to capture the underlying semantics shared across different circuit topologies.

However, using only positive augmentation severely limits the diversity of graph-level functionalities in the pretraining dataset.
For example, if there are 50 distinct original circuits prior to augmentation, positive augmentation will generate new topologies whose graph-level properties remain constrained to these 50 functional types.
As a result, the pretraining dataset would contain graph data representing only 50 unique graph-level functionalities.
To address this, we introduce a negative data augmentation technique designed to increase the functional diversity within the pretraining dataset.
Specifically, we randomly select a device subgraph and modify it into an alternative subgraph, as illustrated in \Cref{fig:dataaugmentation}(a).
This transformation induces significant changes in frequency response or DC voltage behavior, resulting in new topologies with distinct graph-level functionalities (see \Cref{appendix:negda}).

\subsection{Graph Contrastive Pretraining} \label{sec:loss}

Once the augmented pretraining dataset is prepared, graph contrastive learning is performed using graph-level features of the circuits, as illustrated in the upper right part of~\Cref{fig:overview}.

\textbf{Positive, Negative, and Non-Equal Relations.}
We define three types of relationships between circuits in the pretraining dataset: positive, negative, and non-equal.
First, two circuits are in a positive relation if both are generated via positive augmentation from the same original circuit.
Second, two circuits are in a negative relation if one is generated through negative augmentation of the original circuit from which the other is derived.
Third, all remaining pairs fall under the non-equal relation, representing circuits that are augmented from different original circuits.

For example, consider two different original circuits, $x_1$ and $x_2$, prior to augmentation.
Let $x_{1+}$, $x_{2+}$ denote circuits generated via positive augmentation from $x_1$ and $x_2$, respectively.
Also, let $x_{1-}$ and $x_{2-}$ denote circuits generated via negative augmentation.
In this case, positively related pairs include $(x_1, x_{1+})$ and $(x_2, x_{2+})$.
Negatively related pairs include $(x_1, x_{1-})$, $(x_2, x_{2-})$, $(x_{1+}, x_{1-})$, and $(x_{2+}, x_{2-})$.
Non-equal pairs include $(x_1, x_2)$, $(x_1, x_{2+})$, $(x_1, x_{2-})$, $(x_2, x_{1+})$, and $(x_2, x_{1-})$.

Our graph contrastive learning framework maximizes the cosine similarity between positive pairs and minimizes it only between non-equal pairs.
We do not minimize the similarity between negative pairs, as these circuits differ only in localized structures and still share similar overall topologies compared to non-equal pairs.
Empirical results indicate that minimizing the similarity between negative pairs leads to unstable pretraining.
However, negative augmentation still contributes to samples included in non-equal pairs and provide diversity in graph-level functionalities.

\textbf{Model Architecture of DICE.}
We define DICE as a pretrained Graph Isomorphism Network with a depth of 2, incorporating edge feature updates (see \Cref{appendix:dice_gnn_update}).
However, any other GNN architecture can be used.
All feature update rules used in this work are detailed in \Cref{appendix:gnn_updates}.

\textbf{Contrastive Learning Objective.}
We introduce a loss variant of Eq.\eqref{eq:simclr}, formulated in Eq.\eqref{eq:contrastive_learning_loss}.

\begin{equation}
\small
\label{eq:contrastive_learning_loss}
    \mathcal{L}_{DICE}(\theta)=-\underset{x\sim p_{data}}{\mathbb{E}}[\frac{1}{N^+}\underset{x'\in X^+(x)}{\sum}\log\frac{\exp(f_{\theta}(x,x')/\tau)}{\underset{x^+\in X^+(x)}{\sum}\exp(f_{\theta}(x,x^+)/\tau_p)+\underset{x^{\neq}\in X^{\neq}(x)}{\sum}\exp(f_{\theta}(x,x^{\neq})/\tau_n)}]
\end{equation}

The GNN feature extractor $f_{\theta}$, parameterized by $\theta$, computes the cosine similarity between the graph features of two input graphs.
The temperature coefficients are denoted by $\tau$, $\tau_p$, and $\tau_n$, and $p_{\text{data}}$ represents the probability distribution over all circuits in the pretraining dataset.
Let $X^+(x)$ denote the discrete set of samples that are in a positive relation with $x$, and $X^{\neq}(x)$ denote the set of samples that are in a non-equal relation with $x$.
We use $N^+$ to represent the number of samples in the set $X^+(x)$.
Details on the practical implementation of this objective are provided in \Cref{appendix:cl}.

\subsection{Application to Downstream Tasks} \label{sec:dwnstrm}

\begin{figure}[h!]
\begin{center}
\centerline{\includegraphics[width=\columnwidth]{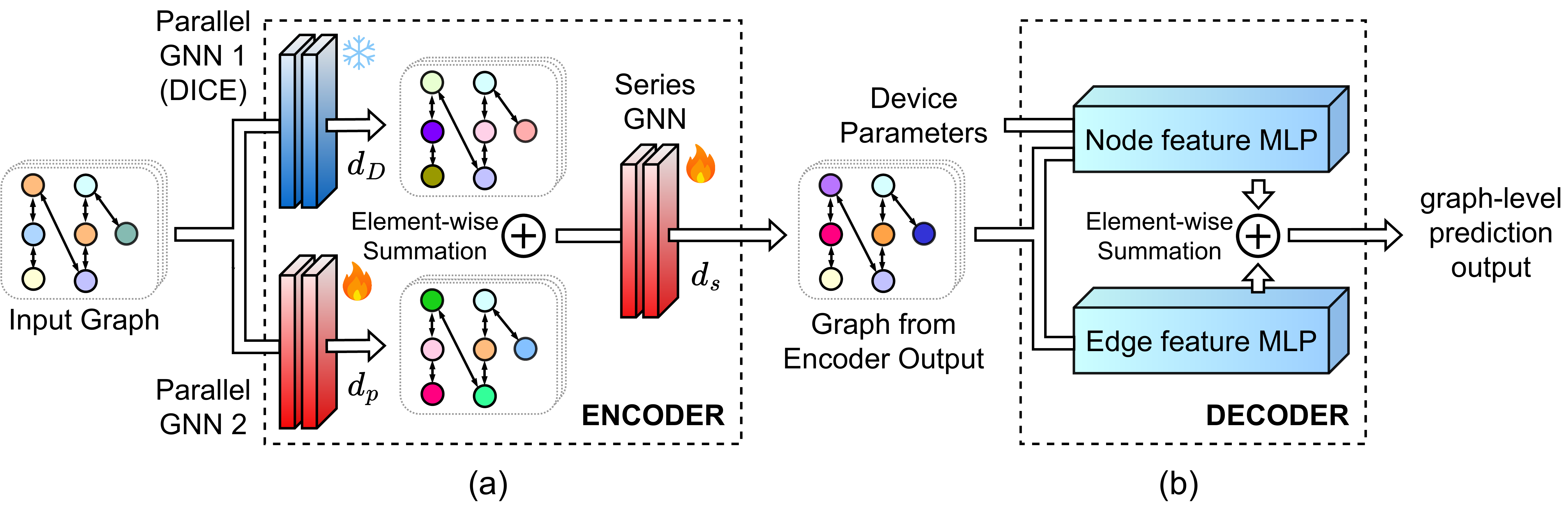}}
\caption{\textbf{Encoder (a)–Decoder (b) model architecture used for solving downstream tasks.} For Parallel GNN 2 and Series GNN, both node and edge features are updated according to Eqs.~\eqref{eq:gnn0}–\eqref{eq:gnn2}.}
\label{fig:fullmodelarch}
\end{center}
\vskip -0.2in
\end{figure}

After pretraining, we integrate DICE into the full message passing neural network (MPNN) model illustrated in \Cref{fig:fullmodelarch} to address diverse downstream tasks.

The encoder (\Cref{fig:fullmodelarch}(a)) comprises three GNN modules: two parallel networks that process the same graph input, followed by a series-connected network.
One of the parallel GNNs is initialized and frozen with the pretrained DICE parameters, while the other parallel GNN and the series-connected GNN remain fully trainable.
The encoder configuration remains fixed for each downstream task if the GNN depths are specified in advance.
We denote the GNN depths of the two parallel networks and the series-connected network as $d_D$, $d_p$, and $d_s$, respectively.

The decoder network (\Cref{fig:fullmodelarch}(b)) takes the output graph from the encoder as input and incorporates device parameters.
To encode these parameters, each device type is represented by a 9-dimensional one-hot vector and scaled by its corresponding parameter values.
Because parameter values can vary significantly in magnitude across device types, we apply a negative logarithmic transformation prior to scaling.
The resulting encoded parameters are then concatenated with the node features from the encoder output and passed through the subsequent layers of the decoder.
\section{Experiments} \label{sec:exp}
We begin by outlining the experimental setup and presenting the results of pretraining.
We then evaluate the pretrained model on three downstream tasks, each formulated as a graph-level prediction problem.
Experimental details are provided in \Cref{appendix:exp}.

\subsection{Pretraining}
\textbf{Setup.}
We compare three different contrastive learning approaches:
The first approach uses $\mathcal{L}_{\text{SimSiam}}$ (Eq.~\eqref{eq:simsiam}) with a pretraining dataset generated solely through positive data augmentation.
The second approach uses $\mathcal{L}_{\text{NT-Xent}}$ (Eq.~\eqref{eq:simclr}), also with a training dataset constructed using only positive augmentation.
The third approach is our proposed method described in \Cref{sec:method}, which uses $\mathcal{L}_{\text{DICE}}$ (Eq.~\eqref{eq:contrastive_learning_loss}) and incorporates both positive and negative data augmentation.
For $\mathcal{L}_{\text{NT-Xent}}$, we set $\tau=0.05$.
For $\mathcal{L}_{\text{DICE}}$, we set $(\tau,\tau_p,\tau_n) = (0.05,0.2,0.05)$.
All GNNs follow Eqs.~\eqref{eq:gnn0}–\eqref{eq:gnn2}.

\begin{figure}[t!]
\begin{center}
\centerline{\includegraphics[width=\columnwidth]{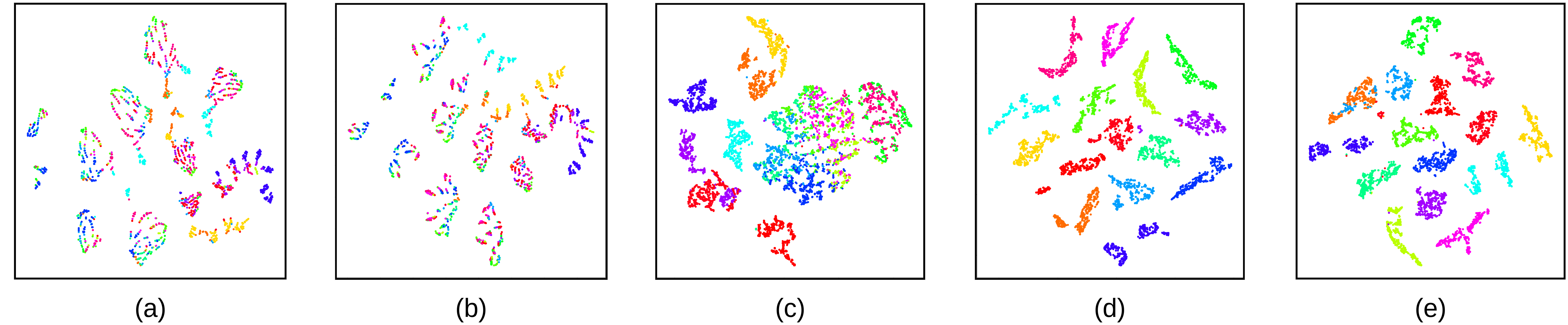}}
\caption{\textbf{t-SNE visualization of graph-level features.} Each plot represents:
(a) initial embeddings,
embeddings processed by an
(b) untrained GNN,
(c) GNN pretrained with $\mathcal{L}_{\text{SimSiam}}$,
(d) GNN pretrained with $\mathcal{L}_{\text{NT-Xent}}$, and
(e) GNN pretrained with $\mathcal{L}_{\text{DICE}}$ (DICE).}
\label{tsne}
\end{center}
\vskip -0.2in
\end{figure}

\textbf{t-SNE Visualization.}
\Cref{tsne} presents the t-SNE plots of graph embedding vectors.
We use the pretraining test dataset (\Cref{appendix:pretraining_dataset}) constructed solely with positive data augmentation, as negatively augmented samples cannot be clearly categorized and may distort the visualization.
Each of the 15 colors represents a distinct initial circuit, and all of its positively augmented views share the same color.
\Cref{tsne}(c), (d), and (e) exhibit improved clustering compared to \Cref{tsne}(a) and (b), providing evidence of effective contrastive pretraining.

\textbf{Cosine Similarity Values.}
To further validate the pretraining results, we computed cosine similarity values between the extracted graph-level features of circuits in the test dataset.
For this analysis, we used the pretraining test dataset (\Cref{appendix:pretraining_dataset}) constructed with both positive and negative data augmentation.
Table~\ref{tab:cosine_similarity} presents the average cosine similarities among pairs in positive, non-equal, and negative relations.
The results using DICE confirm that similarities are high between positively related pairs and effectively minimized between non-equal pairs.
Also, cosine similarity is also minimized between samples in a negative relation, even without direct minimization in Eq.~\eqref{eq:contrastive_learning_loss}.

\begin{table}[h!]
\small
\centering
\caption{\textbf{Cosine similarity values (mean and standard deviation) between graph embeddings in the pretraining test dataset.} With DICE, similarities are minimized between negative pairs.}
\label{tab:cosine_similarity}
\begin{tabular}{cccccc}
\hline\hline
\multirow{2}{*}{Pairs} & \multirow{2}{*}{Initial} & Untrained       & Pretrained GNN                      & Pretrained GNN                      & \multirow{2}{*}{DICE} \\
                       &                          & GNN             & with $\mathcal{L}_{\text{SimSiam}}$ & with $\mathcal{L}_{\text{NT-Xent}}$ &                       \\
\hline\hline
Positive               & $0.998\pm0.002$          & $0.995\pm0.006$ & $0.804\pm0.393$                     & $0.958\pm0.080$                     & $0.906\pm0.161$       \\
Non-equal              & $0.993\pm0.005$          & $0.982\pm0.016$ & $0.124\pm0.708$                     & $0.481\pm0.350$                     & $0.349\pm0.491$       \\
Negative               & $0.997\pm0.002$          & $0.993\pm0.006$ & $0.505\pm0.636$                     & $0.861\pm0.175$                     & $\textbf{0.177}\pm0.362$       \\
\hline\hline
\end{tabular}
\end{table}

\subsection{Downstream Tasks}
To support multiple topologies per task (see \Cref{appendix:prior_graphleveltasks} and \Cref{appendix:prior_benchmarks}), we propose a benchmark consisting of three new graph-level prediction tasks:
(1) predicting the relative similarity among fifty different circuits,
(2) performance prediction for five delay line circuits, and
(3) performance prediction for five operational amplifier circuits.
Detailed descriptions are provided in \Cref{appendix:dwnstrm_settings}.

\textbf{Baseline Setup.}
Our baselines include prior work that (1) explicitly defines graph construction using transistors, (2) employs GNNs, and (3) predicts circuit simulation results~\cite{ren2020paragraph, Hakhamaneshi2023_PretrainGNN}.
Specifically, we selected the unpretrained GNN model (\textbf{ParaGraph}) from \cite{ren2020paragraph}, and both the unpretrained (\textbf{DeepGen$_{u}$}) and pretrained (\textbf{DeepGen$_{p}$}) GNN models from \cite{Hakhamaneshi2023_PretrainGNN} as baselines.
DeepGen$_{p}$ is pretrained on a supervised DC voltage prediction task, as proposed in \cite{Hakhamaneshi2023_PretrainGNN} (see \Cref{appendix:pretrain_deepgen}).
To solve the three downstream tasks, the GNN models from the selected baselines are used as encoders (\Cref{fig:fullmodelarch}(a)), while the decoder component (\Cref{fig:fullmodelarch}(b)) remains identical across all models, including ours.

\begin{figure*}[t!]
\begin{center}
\centerline{\includegraphics[width=\textwidth]{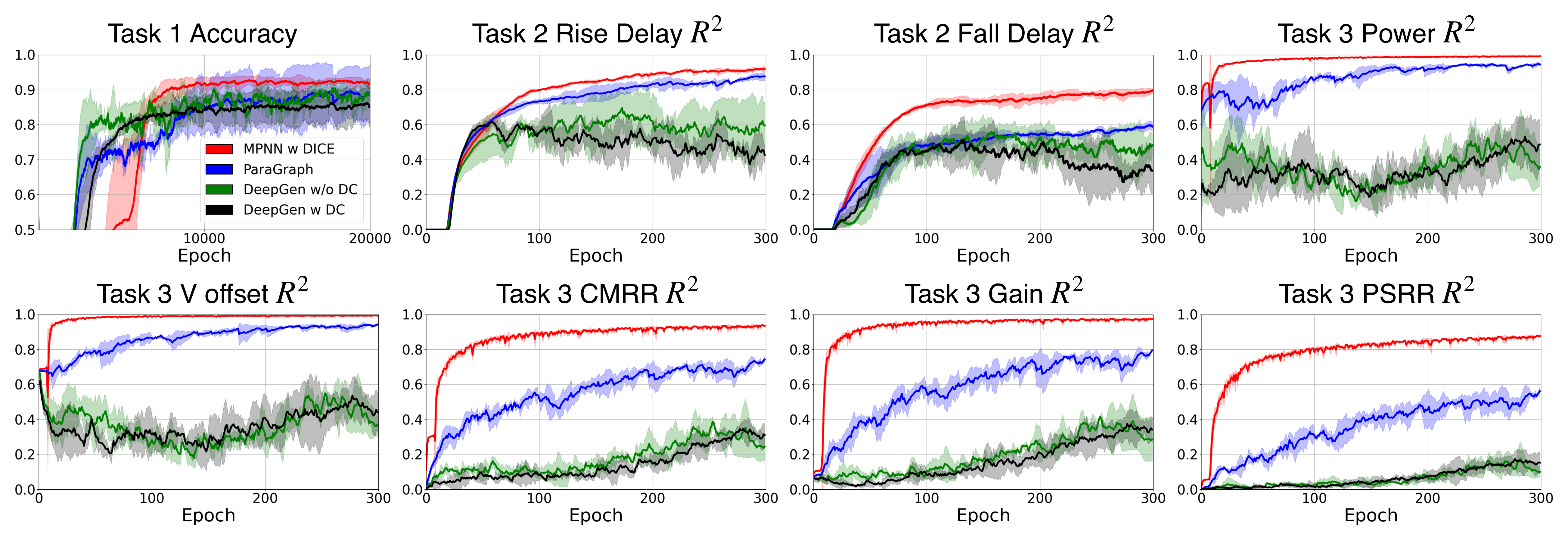}}
\caption{\textbf{Performance curves during downstream task training.}
Each plot shows the $\mu\pm\sigma$ range across 3 different seeds.
We compared our method (red), ParaGraph (blue), DeepGen$_u$ (green), and DeepGen$_p$ (black).
Applying DICE leads to stable training, and consistently outperforms all others.}
\label{fig:baseline_plots}
\end{center}
\vskip -0.2in
\end{figure*}

\textbf{Results.}
We compare the baselines with our encoder model incorporating DICE (\Cref{fig:fullmodelarch}(a)), using a GNN depth configuration of $(d_D, d_p, d_s) = (2, 0, 2)$.
\Cref{fig:baseline_plots} shows the validation performance curves of all encoder-decoder models during training.
Final performance comparisons for each downstream task are summarized in \Cref{tab:experiment1_result}.
Each model was trained using three different random seeds, and we report the average performance across these models.
Both \Cref{fig:baseline_plots} and \Cref{tab:experiment1_result} demonstrate that our model outperforms all baselines across the downstream tasks.

\begin{table*}[t!]
\small
\centering
\caption{\textbf{Baseline performance on the downstream tasks.} For each evaluation metric, the best results are highlighted in bold. The encoder incorporating DICE outperforms all baseline models.}
\label{tab:experiment1_result}
\begin{tabular}{ccccccccc}
\hline\hline
\multicolumn{1}{c|}{Baseline}                                                                    & \multicolumn{1}{c|}{Task 1 (\%)} & \multicolumn{2}{c|}{Task 2 ($R^2$)}          & \multicolumn{5}{c}{Task 3 ($R^2$)}                                                                                                \\ \cline{2-9} 
\multicolumn{1}{c|}{encoders}                                                                                              & \multicolumn{1}{c|}{Accuracy}     & Rise Delay & \multicolumn{1}{c|}{Fall Delay} & Power  & \multicolumn{1}{l}{V$_{\text{offset}}$} & \multicolumn{1}{l}{CMRR} & \multicolumn{1}{l}{Gain} & \multicolumn{1}{l}{PSRR} \\ \hline\hline
\multicolumn{1}{c|}{ParaGraph}                                                                       & \multicolumn{1}{c|}{91.42}        & 0.9255     & \multicolumn{1}{c|}{0.6373}     & 0.9312 & 0.9585                                  & 0.7912                   & 0.8488                   & 0.6215                   \\
\multicolumn{1}{c|}{\begin{tabular}[c]{@{}c@{}}DeepGen$_u$\end{tabular}} & \multicolumn{1}{c|}{85.02}        & 0.8866     & \multicolumn{1}{c|}{0.6715}     & 0.8403 & 0.8478                                  & 0.5637                   & 0.6813                   & 0.3820                   \\ 
\multicolumn{1}{c|}{\begin{tabular}[c]{@{}c@{}}DeepGen$_p$\end{tabular}}    & \multicolumn{1}{c|}{84.65}        & 0.9131     & \multicolumn{1}{c|}{0.6875}     & 0.4479 & 0.5190                                  & 0.3246                   & 0.3454                   & 0.0851                   \\ \hline\hline
\multicolumn{1}{c|}{DICE}                                                                              & \multicolumn{1}{c|}{\textbf{93.94}}        & \textbf{0.9468}     & \multicolumn{1}{c|}{\textbf{0.8085}}     & \textbf{0.9916} & \textbf{0.9950}                                  & \textbf{0.9374}                   & \textbf{0.9769}                   & \textbf{0.8809}                   \\ \hline\hline
\end{tabular}
\vskip -0.1in
\end{table*}

\subsection{Ablation Studies}
\textbf{(1) $\mathcal{L}_{\text{SimSiam}}$ vs. $\mathcal{L}_{\text{NT-Xent}}$ vs. $\mathcal{L}_{\text{DICE}}$.}
We also applied GNNs pretrained with $\mathcal{L}_{\text{SimSiam}}$ and $\mathcal{L}_{\text{NT-Xent}}$ in the encoder of \Cref{fig:fullmodelarch} and evaluated them on our downstream tasks.
As shown in \Cref{tab:ablation_1_2}, both methods improve performance over the baselines in tasks 2 and 3.

\textbf{(2) DICE vs. Diverse GNNs.}
We also evaluated the effect of pretraining different GNN architectures, with results provided in \Cref{tab:ablation_1_2}.
GCN~\cite{gcn}, GraphSAGE~\cite{graphsage}, GAT~\cite{gat}, and GIN~\cite{gin} were pretrained using $\mathcal{L}_{\text{DICE}}$ (see \Cref{appendix:gnn_updates}).
Pretraining diverse GNNs also leads to performance gains.

\begin{table}[b!]
\centering
\scriptsize
\vskip -0.1in
\caption{\textbf{Results of ablation studies (1) and (2).} Bolded values outperform all baselines in \Cref{tab:experiment1_result}.
Each pretrained GNN serves as part of the encoder in \Cref{fig:fullmodelarch}, with $(d_D, d_p, d_s) = (2, 0, 2)$.}
\label{tab:ablation_1_2}
\begin{tabular}{cc|c|cc|ccccc}
\hline\hline
Pretraining                    & Pretrained & Task 1 (\%) & \multicolumn{2}{c|}{Task 2 ($R^2$)} & \multicolumn{5}{c}{Task 3 ($R^2$)}               \\ \cline{3-10} 
Loss                           & GNN        & Accuracy         & Rise Delay       & Fall Delay       & Power           & V$_{offset}$     & CMRR   & Gain   & PSRR   \\ \hline\hline
$\mathcal{L}_{\text{SimSiam}}$ & DICE       & 85.88            & \textbf{0.9427}  & \textbf{0.8044}  & \textbf{0.9911} & \textbf{0.9951}  & \textbf{0.9353} & \textbf{0.9762} & \textbf{0.8768} \\
$\mathcal{L}_{\text{NT-Xent}}$ & DICE       & 76.73            & \textbf{0.9492}  & \textbf{0.7706}  & \textbf{0.9898} & \textbf{0.9947}  & \textbf{0.9328} & \textbf{0.9757} & \textbf{0.8720} \\ \hline\hline
$\mathcal{L}_{\text{DICE}}$    & GCN        & 85.92            & 0.9113           & \textbf{0.7467}  & \textbf{0.9909} & \textbf{0.9940}  & \textbf{0.9314} & \textbf{0.9743} & \textbf{0.8662} \\
$\mathcal{L}_{\text{DICE}}$    & GraphSAGE  & 79.35            & 0.8975           & \textbf{0.7786}  & \textbf{0.9899} & \textbf{0.9947}  & \textbf{0.9346} & \textbf{0.9756} & \textbf{0.8702} \\
$\mathcal{L}_{\text{DICE}}$    & GAT        & 84.32            & \textbf{0.9314}  & \textbf{0.7981}  & \textbf{0.9862} & \textbf{0.9867}  & \textbf{0.9130} & \textbf{0.9624} & \textbf{0.8004} \\
$\mathcal{L}_{\text{DICE}}$    & GIN        & \textbf{93.88}   & \textbf{0.9414}  & \textbf{0.7931}  & \textbf{0.9915} & \textbf{0.9947}  & \textbf{0.9336} & \textbf{0.9757} & \textbf{0.8788} \\ \hline\hline
$\mathcal{L}_{\text{DICE}}$    & DICE       & \textbf{93.94}   & \textbf{0.9468}  & \textbf{0.8085}  & \textbf{0.9916} & \textbf{0.9950}  & \textbf{0.9374} & \textbf{0.9769} & \textbf{0.8809} \\ \hline\hline
\end{tabular}
\end{table}

\textbf{(3) Training from Scratch vs. Applying DICE.}
We compared two models: one trained from scratch and the other incorporating DICE.
With both models following the architecture shown in \Cref{fig:fullmodelarch}, the encoder remains fixed across all downstream tasks, and the decoder varies by task but is consistent across models within each task.
$d_D$ is set to 0 for models without DICE, and $d_D$ is set to 2 for models with DICE.
The number of trainable parameters is kept constant across models with identical $(d_p, d_s)$ values.
\Cref{tab:ablation3_result} presents the experimental results, and all bolded and underlined values correspond to the MPNN model incorporating DICE, except for a few cases when $(d_p, d_s) = (1,1)$.

\begin{table}[t!]
\centering
\small
\caption{\textbf{Results on ablation study (3).}
The final performance values are bolded if they surpass the value of the counterparts, which indicates models having the same ($d_p$, $d_s$) values but different $d_D$ values.
Per task metric, the best performance values across all models are marked with underlines.}
\label{tab:ablation3_result}
\begin{tabular}{cc|c|cc|ccccc}
\hline\hline
\multicolumn{2}{r|}
{\multirow{2}{*}
{($d_D,d_p,d_s$)}}      & Task 1 (\%) & \multicolumn{2}{c|}{Task 2 ($R^2$)} & \multicolumn{5}{c}{Task 3 ($R^2$)}        \\ \cline{3-10} 
\multicolumn{2}{r|}{}   & Accuracy & $d_{rise}$ & $d_{fall}$ & Power & V$_{offset}$ & CMRR & Gain & PSRR      \\ \hline\hline
\multirow{6}{*}
{\begin{tabular}[c]
{@{}c@{}}
MPNN\\without\\DICE
\end{tabular}}          & (0, 0, 0) & 76.91 & 0.7541 & 0.5456 & 0.9651 & 0.8261 & 0.4794 & 0.3754 & 0.2798 \\
                        & (0, 0, 1) & 74.84 & 0.7706 & 0.5486 & 0.9718 & 0.8976 & 0.6384 & 0.6441 & 0.5024 \\
                        & (0, 1, 0) & 77.90 & 0.7666 & 0.5405 & 0.9636 & 0.8825 & 0.6019 & 0.6013 & 0.4501 \\
                        & (0, 1, 1) & 84.68 & 0.9404 & \textbf{0.8124} & 0.9827 & \textbf{0.9876} & 0.9125 & \textbf{0.9633} & \textbf{0.8064} \\
                        & (0, 0, 2) & 88.57 & 0.9229 & 0.7790 & 0.9745 & 0.9863 & 0.9091 & 0.9605 & 0.7929 \\
                        & (0, 2, 0) & 86.28 & 0.9237 & 0.7618 & 0.9866 & 0.9867 & 0.9066 & 0.9610 & 0.7900 \\ \hline\hline
\multirow{6}{*}
{\begin{tabular}[c]
{@{}c@{}}
MPNN\\with\\DICE
\end{tabular}}          & (2, 0, 0) & \textbf{92.13} & \textbf{0.9352} & \textbf{0.7885} & \textbf{0.9769} & \textbf{0.9820} & \textbf{0.8865} & \textbf{0.9402} & \textbf{0.7535} \\
                        & (2, 0, 1) & \textbf{93.41} & \textbf{0.9310} & \textbf{0.7805} & \textbf{0.9904} & \textbf{0.9876} & \textbf{0.9167} & \textbf{0.9659} & \textbf{0.8132} \\
                        & (2, 1, 0) & \textbf{93.43} & \textbf{0.9294} & \textbf{0.7820} & \textbf{0.9849} & \textbf{0.9868} & \textbf{0.9147} & \textbf{0.9634} & \textbf{0.8061} \\
                        & (2, 1, 1) & \textbf{\underline{94.54}} & \textbf{\underline{0.9417}} & 0.7873          & \textbf{0.9847} & 0.9872 & \textbf{0.9147} & 0.9627 & 0.8061 \\
                        & (2, 0, 2) & \textbf{94.01} & \textbf{0.9231} & \textbf{\underline{0.8134}} & \textbf{\underline{0.9918}} & \textbf{\underline{0.9950}} & \textbf{\underline{0.9371}} & \textbf{\underline{0.9769}} & \textbf{\underline{0.8776}} \\
                        & (2, 2, 0) & \textbf{93.98} & \textbf{0.9402} & \textbf{0.7851} & \textbf{0.9876} & \textbf{0.9874} & \textbf{0.9165} & \textbf{0.9641} & \textbf{0.8055} \\ 
\hline\hline
\end{tabular}
\vskip -0.2in
\end{table}
\section{Discussion} \label{sec:discussion}

Based on the results from both baseline and ablation studies, we highlight four key observations:

First, applying pretrained GNNs significantly improves performance compared to training from scratch.
As shown in \Cref{tab:experiment1_result}, our model using DICE outperforms ParaGraph with up to a 15.19\% average improvement on Task 3, and exceeds DeepGen$_u$ with up to a 44.24\% average improvement on the same task.
Ablation study (3) further confirms the effectiveness of DICE, achieving up to a 19.79\% improvement on Task 1 when $(d_p, d_s) = (0, 0)$.

Second, graph contrastive pretraining is substantially more effective than the supervised pretraining method proposed in~\cite{Hakhamaneshi2023_PretrainGNN}.
In \Cref{tab:experiment1_result}, our model outperforms DeepGen$_p$ by 10.97\% on Task 1 and achieves a 9.67\% average improvement on Task 2.
These results underscore the advantage of unsupervised learning over supervised approaches for task-agnostic pretraining.

Third, combining negative augmentation with similarity minimization improves performance.
In Task 1 of ablation study (1), our model incorporating DICE outperforms the one pretrained with $\mathcal{L}_{\text{SimSiam}}$ by 9.39\%, and exceeds the model pretrained with $\mathcal{L}_{\text{NT-Xent}}$ by 22.43\%.
The results follow the conclusion in \cite{chen2020simple}, arguing that similarity minimization is not necessary in contrastive learning. However, results also show that using $\mathcal{L}_{\text{DICE}}$ with our negative augmentation outperforms the approach in \cite{chen2020simple}.

Fourth, the proposed pretraining framework generalizes well across diverse GNN architectures.
In Task 3 of ablation study (2), all four pretrained models outperform ParaGraph by an average of 14.09\% and exceed DeepGen$_u$ by 42.87\%.
These results demonstrate the robustness of our method.
\section{Conclusion} \label{sec:conclu}

In this work, we propose DICE, a pretrained GNN model designed for general graph-level prediction tasks in device-level integrated circuits.
Our primary contribution is to highlight the importance of pretraining graph models specifically on device-level circuits, and we introduce the first graph contrastive learning framework to address this challenge.
We argue that device-level representations offer a more general and flexible abstraction for integrated circuits than logic gate-level representations, as they support both analog and digital designs.
Another contribution is the introduction of two novel data augmentation techniques that address the scarcity of circuit data.
Additionally, we propose a new device-level circuit benchmark, where all three tasks require handling multiple circuit topologies.
Experimental results including the ablation studies show that incorporating DICE leads to significant improvements across all three downstream tasks.
We view this work as a step toward building general-purpose models that unify analog and digital circuit design.

\bibliographystyle{unsrtnat}
\bibliography{ref/reference}

\appendix
\newpage
\appendix
\onecolumn

\section{Related Works}
\label{appendix:related_works}

\subsection{Graph-level Prediction Tasks in EDA}
\label{appendix:prior_graphleveltasks}

Predicting properties of the entire circuit can significantly reduce design time, as simulations are often time-consuming.
To address this, prior work on digital circuits has focused on training models to predict power, performance, and area (PPA)~\cite{Li2022NoCeption, Fang2023MasterRTL, Du2024PowPrediCT}, timing violations~\cite{Guo2022TimingEngineGNN, Gandham2024CircuitSeer}, routing congestions~\cite{Wang2022LHNN, Zhao2024PDNNet}, and functional similarities between digital circuits~\cite{Alrahis2022GNN-RE, bucher2022appgnn}.
For analog circuits, previous studies have trained models to predict performance metrics such as gain, bandwidth, and power consumption~\cite{ren2020paragraph, Hakhamaneshi2023_PretrainGNN, Shahane2023_GCX, Wu2023_CircuitGNN, Khamis2024_DCDC_GNN, Liu2021_ParasiticGNN, Chen2021_SymmetryGNN, chae2024pulserf}, and also trained models to identify functional similarities between analog circuit blocks or transistor groups~\cite{deeb2023robust, deeb2024graph, kunal2023gnn, xu2024graph}.
In this work, we show that our pretrained model DICE is effective in solving similar graph-level prediction tasks.

\subsection{Benchmarks for Device-level Circuits}
\label{appendix:prior_benchmarks}
Benchmarks for graph-level prediction tasks exist for both analog and digital circuits; however, digital circuit benchmarks are unsuitable for device-level circuits because their graph representations are based on logic gates rather than device components.
Analog circuit benchmarks also pose challenges when used to evaluate graph learning methods at the device level.
The Open Circuit Benchmark (OCB)\cite{dong2023cktgnn} provides various operational amplifier topologies, but it is not suitable for device-level circuit evaluation since its circuit graphs do not use transistors as fundamental components.
Analoggym\cite{li2024analoggym} and AICircuit~\cite{mehradfar2024aicircuit} offer multiple circuit topologies, but the associated tasks focus on device parameter optimization, which are inherently node-level tasks.
While \cite{poddar2024insight, wang2024learn} address graph-level prediction, each task in their benchmarks involves only a single circuit topology.
These limitations underscore the need for new graph-level prediction benchmarks that support multiple circuit topologies and represent circuits using device-level components such as transistors.

\subsection{Converting Device-Level Circuits into Graphs}
\label{appendix:motiv_graph_construct}

There exist various approaches for converting device-level circuits into graphs.
A key challenge lies in how to represent transistor devices as graph components.
The work in~\cite{ren2020paragraph} models both devices and voltage nets as graph nodes and distinguishes connection types using multiple edge types.
They adopt a heterogeneous GNN (ParaGraph) with an attention mechanism, and omit power and ground voltage nets from the graph as nodes.
The studies in~\cite{hakhamaneshi2022pretraining, Hakhamaneshi2023_PretrainGNN} pretrain the GNN (DeepGen) proposed in~\cite{li2020deepergcn}, using a supervised node-level task: DC voltage prediction.
Their graph construction does not differentiate edge types and instead separates device terminals into individual nodes, disjoint from their corresponding device nodes.

Recent works~\cite{hakhamaneshi2022pretraining, Hakhamaneshi2023_PretrainGNN, gao2025analoggenie} follow a similar modeling approach, treating each device pin as a separate node.
However, such representations can lead to inefficient message passing when applied in Message Passing Neural Networks (MPNNs).
For example, the edge from a gate node to a transistor device should be directional, not bidirectional.
This is because the voltage signal at the gate controls the current flow through the transistor, but the transistor exerts minimal influence on the gate voltage.
To address this limitation, we introduce a new mapping method that maintains directional semantics with edge features and minimizes redundant message passing.

\subsection{Including Device Parameters in the Pretraining Dataset}
\label{appendix:motiv_data_aug}
Training a general-purpose model requires constructing a comprehensive dataset.
A straightforward approach to building a large-scale circuit dataset is to enumerate various device parameter configurations~\cite{dong2023cktgnn, mehradfar2024aicircuit}.
However, this method may fail to capture the full breadth of circuit knowledge.
Consider a circuit with twenty devices, each with three possible parameter values—this results in $3^{20}$ potential combinations, yielding an enormous dataset from a single topology.
Yet, constraining each device to only three parameter values is insufficient to represent the true complexity of circuit behavior, as meaningful parameters vary widely across different tasks and technologies.
Moreover, capturing structural diversity requires incorporating multiple circuit topologies, which further expands the dataset and introduces additional inefficiencies.
Therefore, instead of relying on device parameter variations, we employ data augmentation to generate a sufficiently large pretraining dataset.

\newpage

\section{Details of Data Augmentation}
\label{appendix:da}

\subsection{Positive Data Augmentation Details}
\label{appendix:posda}

We explain why positive data augmentation leads to the preservation of graph-level functionality.
Consider two resistors with resistance values $r_1$ and $r_2$.
When connected in series, these resistors can be equivalently replaced by a single resistor with resistance value of $r_1+r_2$.
Similarly, for parallel connections, they can be replaced by a resistor with resistance value of $\frac{r_1r_2}{r_1+r_2}$.
This principle applies analogously to all other circuit components.
Since device parameters do not affect structural circuit properties, we can consolidate series/parallel-connected devices into single components.
Consequently, we can make positive samples consistently converge toward unified topological structures (in our case, 55 initial circuits) through these replacements.

\begin{figure}[h!]
    \begin{center}
    \centerline{\includegraphics[width=0.7\linewidth]{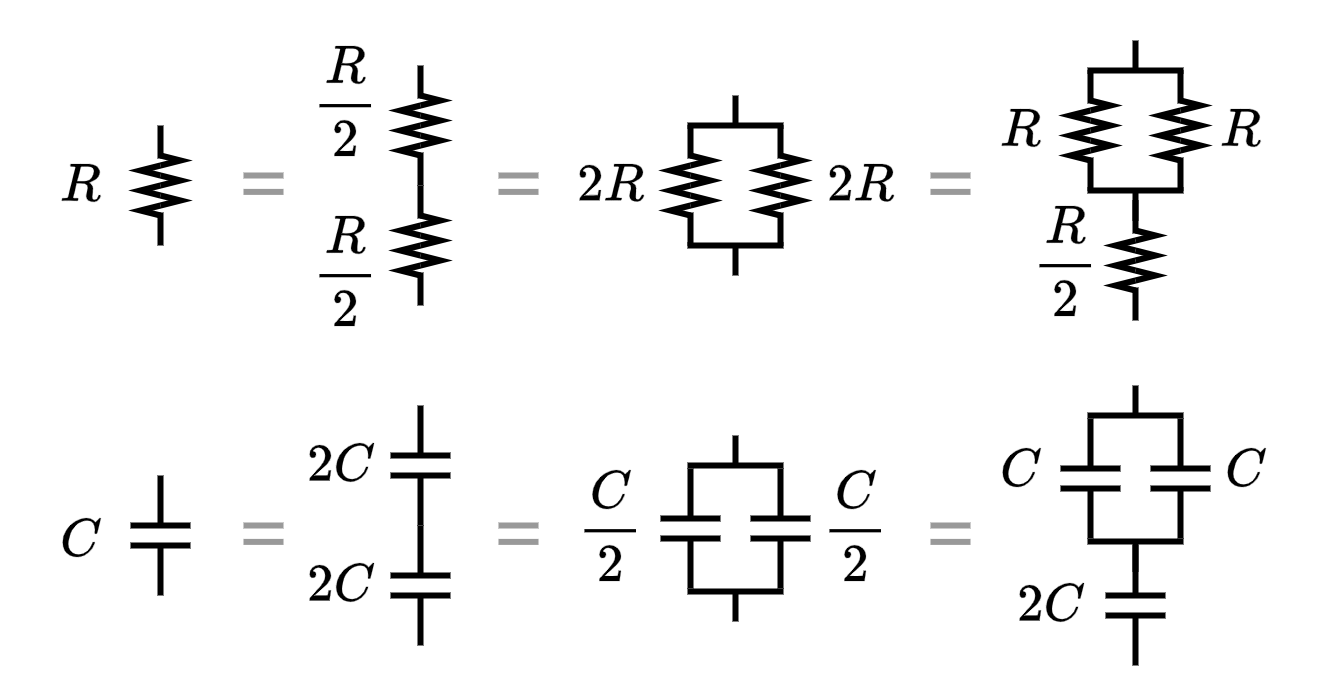}}
    \caption{Equivalent subgraphs generated through positive data augmentation. Different subgraph topologies indicate the same local functionality, and this leads to equal graph-level functions if all other structures are the same.}
    \label{dadetail}
    \end{center}
\vskip -0.2in
\end{figure}

\newpage
\subsection{Negative Data Augmentation Details}
\label{appendix:negda}

We explain why negative data augmentation leads to the perturbation of graph-level functionality.
For passive elements, we consider their impedance values at both DC (zero frequency) and the high-frequency limit, and modify them to have the inverted impedance characteristics at both extremes.
For current sources, we replace them with passive elements, which are equivalent to converting the energy supplier to an energy dissipator.
Finally, for transistors, we consider switching behaviors based on gate voltage inputs and connect the counterpart transistors in both parallel and series.
Since either impedance characteristics, energy consumption, or switching behavior is inverted, the graph-level functionality is modified with our proposing negative data augmentation.

\begin{figure}[h!]
    \begin{center}
    \centerline{\includegraphics[width=\linewidth]{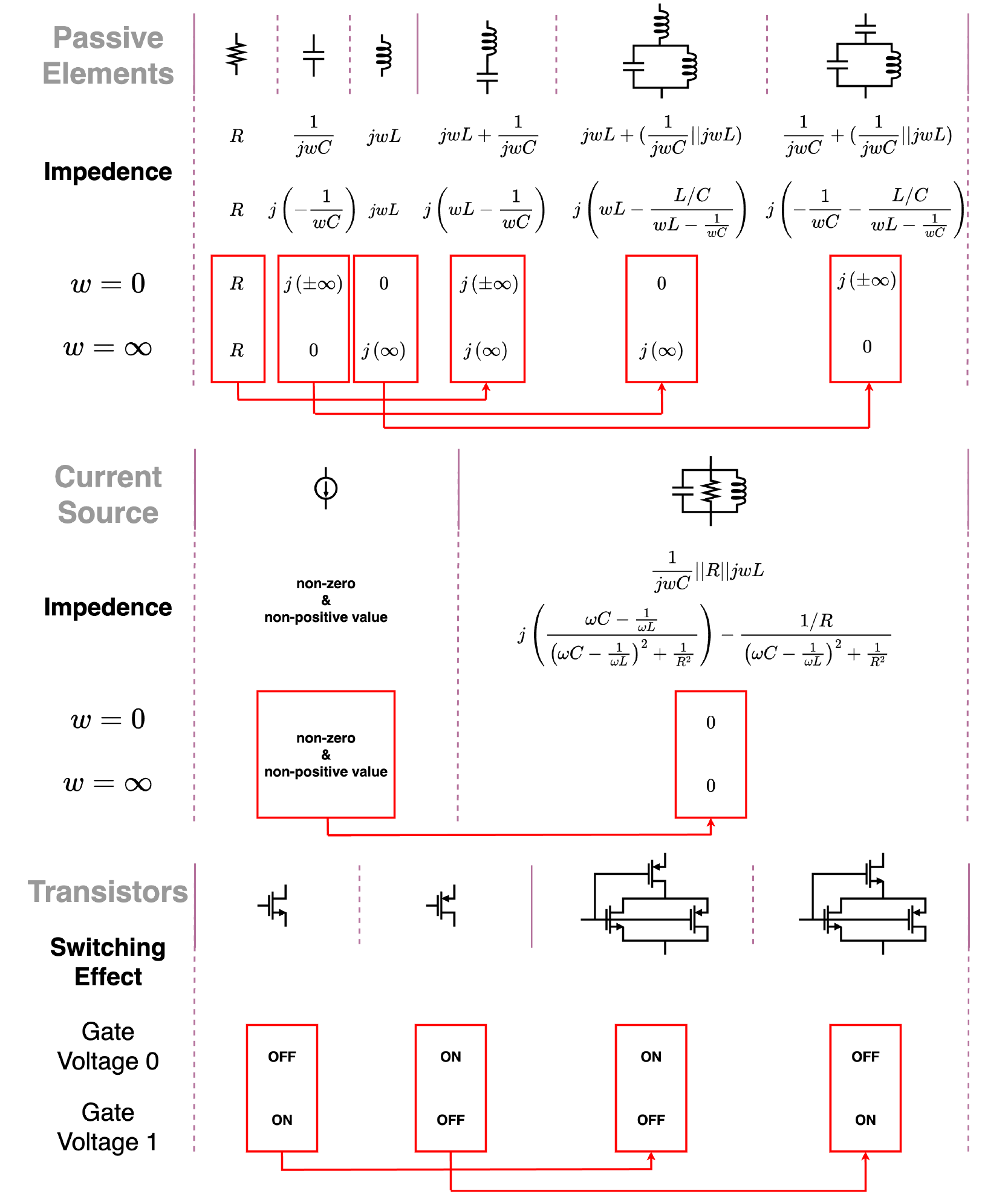}}
    \caption{Conversion of characteristics through negative data augmentation.}
    \label{dadetail}
    \end{center}
\vskip -0.2in
\end{figure}

\section{GNN update rules}
\label{appendix:gnn_updates}

\subsection{DICE}
\label{appendix:dice_gnn_update}
The update rule of DICE follows the Graph Isomorphism Network (GIN) with edge feature updates.

\begin{figure}[H]
  \small
  \begin{equation}
    \label{eq:gnn0}
    \begin{split}
    \mathbf{h}_v^{(1)} &= MLP_{\theta_0}(\mathbf{h}_v^{(0)}), \quad \mathbf{e}_{u\rightarrow v}^{(1)} = MLP_{\theta_1}(\mathbf{e}_{u\rightarrow v}^{(0)})
    \end{split}
  \end{equation}
  \begin{equation}
    \label{eq:gnn1}
    \begin{split}
    \mathbf{m}_v^{(k)} &= \sum_{u\in \mathcal{N}(v)}\mathbf{h}_v^{(k)}\cdot \mathbf{e}_{u\rightarrow v}^{(k)}, \\
    \mathbf{h}_v^{(k+1)} &= MLP_{\theta_{2}^{(k)}}((1+\phi_{h}^{(k)})\cdot \mathbf{h}_v^{(k)} + \mathbf{m}_v^{(k)}), \\
    \mathbf{e}_{u\rightarrow v}^{(k+1)} &= MLP_{\theta_{3}^{(k)}}((1+\phi_{e}^{(k)})\cdot \mathbf{e}_{u\rightarrow v}^{(k)} + \mathbf{m}_u^{(k)} - \mathbf{m}_v^{(k)}) \\
    \end{split}
  \end{equation}
  \begin{equation}
    \label{eq:gnn2}
    \begin{split}
    \mathbf{g}_{(V,E)} &= \sum_{v\in V}\mathbf{h}_v^{(l+1)}+\sum_{u\rightarrow v \in E}\mathbf{e}_{u\rightarrow v}^{(l+1)}
    \end{split}
  \end{equation}
\vskip -0.2in
\end{figure}

Eqs.~\eqref{eq:gnn0}–\eqref{eq:gnn2} illustrates the node and edge feature update rules of DICE, where $\theta_0$, $\theta_1$, $\theta_2^{(k)}$, $\theta_3^{(k)}$, $\phi_h^{(k)}$, and $\phi_e^{(k)}$ are the training parameters. Batch normalization and dropout layers are included between linear layers. Eq.~\eqref{eq:gnn0} matches the initial dimensions of node and edge features (9 and 5) to the hidden dimension value, and $\mathbf{h}_u^{(0)}$, $\mathbf{e}_{v\rightarrow u}^{(0)}$ are the initial one-hot encoding of node and edge types. Eq.~\eqref{eq:gnn1} explains the message passing rule, where $k=1,...,l$ and $l(=2)$ is the depth of the GNN. The graph-level feature $\mathbf{g}_{(V,E)}$ is calculated in Eq.~\eqref{eq:gnn2}, by summing all the updated node and edge features within each graph $G=(V,E)$.

\subsection{GCN}
\label{appendix:gcn_update}
The update rule of the Graph Convolutional Network (GCN) used in this work follows a normalized aggregation mechanism:

\begin{equation}
\begin{split}
\mathbf{m}_v^{(k)} &= \sum_{u \in \mathcal{N}(v)} \frac{1}{\sqrt{d_v d_u}} \cdot \mathbf{h}_u^{(k)} \\
\mathbf{h}_v^{(k+1)} &= \text{MLP}_{\theta^{(k)}}(\mathbf{m}_v^{(k)})
\end{split}
\end{equation}

Here, $d_v$ and $d_u$ denote the degrees of nodes $v$ and $u$, respectively. The GCN layer performs degree-normalized aggregation followed by a learnable transformation of the aggregated features. Although edge features are not explicitly involved in the message-passing process for GCN, Eq.~\eqref{eq:gnn0} is still applied at the initial layer to project both node and edge features into a common hidden space. Eq.~\eqref{eq:gnn2} is used to compute the graph-level feature by summing the final node and edge embeddings.

\subsection{GraphSAGE}
\label{appendix:graphsage_update}
GraphSAGE in our work performs aggregation by concatenating the central node’s embedding with the mean of its neighbors’ embeddings:

\begin{equation}
\begin{split}
\mathbf{m}_v^{(k)} &= \frac{1}{|\mathcal{N}(v)|} \sum_{u \in \mathcal{N}(v)} \mathbf{h}_u^{(k)} \\
\mathbf{h}_v^{(k+1)} &= \text{MLP}_{\theta^{(k)}}\left([\mathbf{h}_v^{(k)} \, || \, \mathbf{m}_v^{(k)}] \right)
\end{split}
\end{equation}

Aggregation is performed through mean pooling, and node features are updated by concatenating the aggregated messages with the node’s current embedding. Although edge features do not contribute to the message-passing step, Eq.~\eqref{eq:gnn0} is used to initialize the edge feature embeddings. The final graph-level feature is computed via Eq.~\eqref{eq:gnn2}, which includes both node and edge representations.

\subsection{GAT}
\label{appendix:gat_update}
The Graph Attention Network (GAT) used in our work computes attention coefficients for each edge to weigh the contributions of neighboring nodes:

\begin{equation}
\begin{split}
\alpha_{u \rightarrow v}^{(k)} &= \text{softmax}_{u \in \mathcal{N}(v)} \left( \mathbf{h}_u^{(k)} \cdot \mathbf{h}_v^{(k)} \right) \\
\mathbf{m}_v^{(k)} &= \sum_{u \in \mathcal{N}(v)} \alpha_{u \rightarrow v}^{(k)} \cdot \mathbf{h}_u^{(k)} \\
\mathbf{h}_v^{(k+1)} &= \mathbf{h}_v^{(k)} + \mathbf{m}_v^{(k)}
\end{split}
\end{equation}

The attention mechanism allows the model to learn the relative importance of neighboring nodes based on their feature similarity. While edge features are not updated during message passing in GAT, they are initially transformed using Eq.~\eqref{eq:gnn0}. The final graph-level representation is computed by summing both the node and edge embeddings as described in Eq.~\eqref{eq:gnn2}.

\subsection{GIN}
\label{appendix:gin_update}
The update rule of the Graph Isomorphism Network (GIN) uses sum aggregation followed by a learnable MLP:

\begin{equation}
\begin{split}
\mathbf{m}_v^{(k)} &= \sum_{u \in \mathcal{N}(v)} \mathbf{h}_u^{(k)} \\
\mathbf{h}_v^{(k+1)} &= \text{MLP}_{\theta^{(k)}}\left((1 + \epsilon^{(k)}) \cdot \mathbf{h}_v^{(k)} + \mathbf{m}_v^{(k)} \right)
\end{split}
\end{equation}

Here, $\epsilon^{(k)}$ is a learnable scalar, initialized as zero. GIN is known for its strong expressive power in distinguishing graph structures. Although the edge features are not used in message passing for GIN, they are projected to the hidden space using Eq.~\eqref{eq:gnn0}. Both node and edge embeddings are summed to form the graph-level representation, as defined in Eq.~\eqref{eq:gnn2}.

\newpage
\section{Details of the Experiments}
\label{appendix:exp}

\subsection{Pretraining}
\label{appendix:pretraining_dataset}
\subsubsection{Dataset}
Data augmentation is used exclusively for generating the pretraining datasets and is performed independently for the pretraining training and test sets.
The pretraining dataset comprises 55 initial circuit topologies, covering both analog and digital designs.
For the training set, 40 topologies are each augmented 4,000 times (2,000 positive and 2,000 negative), resulting in a total of 160,000 graphs.
For testing with t-SNE visualization, the remaining 15 topologies are each augmented 500 times using only positive augmentation, yielding 7,500 graphs.
For testing cosine similarity values, the same 15 topologies are each augmented 500 times (250 positive and 250 negative), also resulting in 7,500 graphs.
We did not use negative augmentation for t-SNE visualization since samples with negative augmentation cannot be categorized and distracts the visualization results.

\subsubsection{Contrastive Learning}
\label{appendix:cl}

\begin{figure*}[h]
    \centering
    \includegraphics[width=0.8\linewidth]{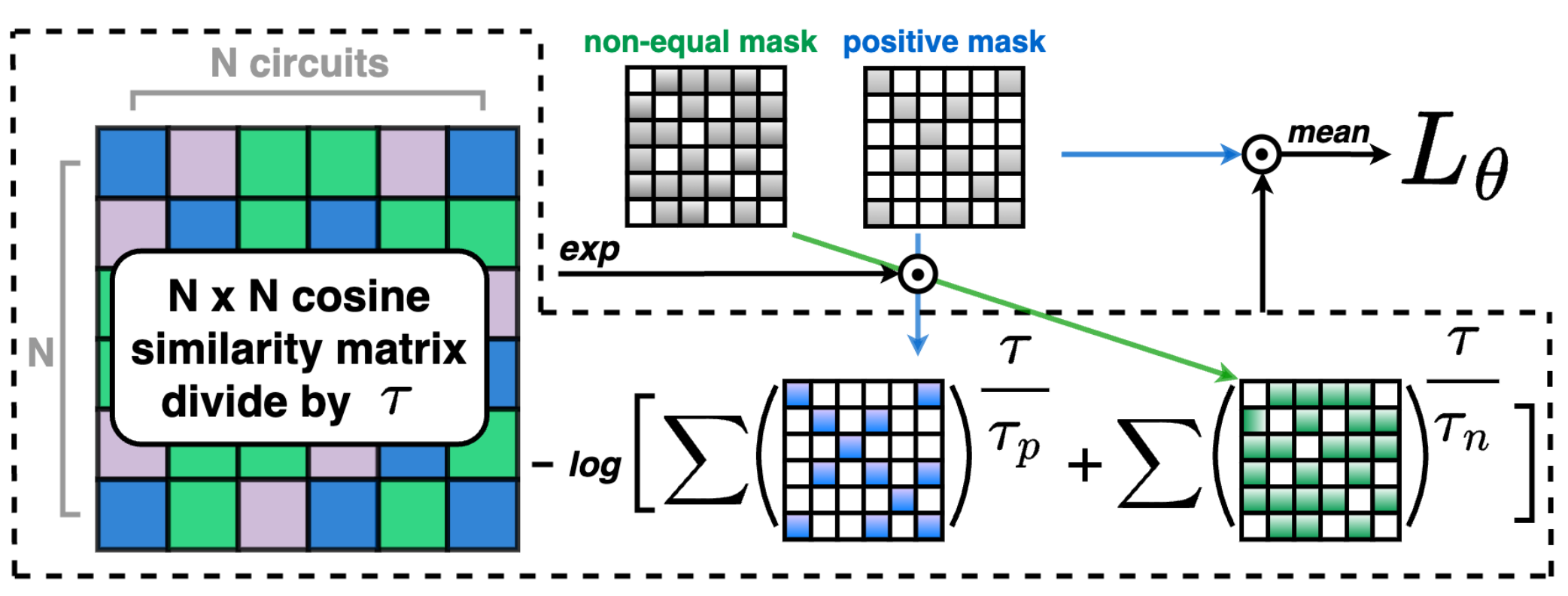}
    \caption{\textbf{Masking operation.} $\odot$ indicates the hadamard product (element-wise multiplication).}
    \label{fig:cldetail}
\end{figure*}

Masking is our key source of rapid pretraining, with the cost of GPU memory. Algorithm.~\ref{alg:contrastive_learning} provides the pseudocode and Fig.~\ref{fig:cldetail} visualize the details of the the proposed masking for contrastive learning. For each batch, we first matched the number of graphs for every circuit types. For example, consider the case when the number of graphs originated from the inverter circuit is 5. If this is the minimum number among all other circuit types in the batch, then 5 circuits are sampled for every circuit types and form a new batch. Then, the cosine similarity matrix is formed with graph-level features in the batch with masks indicating positive and non-equal. Element-wise products between the similarity matrix and the masks filter out the unnecessary parts in parallel. Multiple positive pairs are considered simultaneously with this operation, thereby significantly reducing the training speed with the cost of GPU memory.

\DeclareRobustCommand{\rchi}{{\mathpalette\irchi\relax}}
\newcommand{\irchi}[2]{\raisebox{\depth}{$#1\chi$}} 
\begin{algorithm}[h]
\caption{Graph-level Contrastive Learning}\label{alg:contrastive_learning}
\centering
\begin{minipage}{\columnwidth}
\textbf{Input:} augmented dataset ($\rchi$), epoch ($T$), temperature coefficients ($\tau_p$, $\tau$, $\tau_n$), learning rate ($\eta$) \\
\textbf{Output:} pretrained model parameter ($\theta_T$) \\
\textbf{Initialize:} model parameter $\theta_0$ \\
1:\ \textbf{for} $t=0,1,2,...,T-1$ \textbf{do}\\
2:\quad \textbf{for} batch $B$ in $\rchi$ \textbf{do} \\
3:\quad\quad       $m$                    $\gets$ min number covering all circuit types in $B$\\
4:\quad\quad       $B_m$                  $\gets$ batch$\in \mathbb{R}^L$ having $m$ graphs each for every circuit types (maintaining circuit diversity)\\
5:\quad\quad       $g$                    $\gets$ $GNN_{\theta_t}(B_m)$: graph-level features $\in \mathbb{R}^{L\times h}$ \\
6:\quad\quad       $S$                    $\gets$ $gg^T$: cosine similarity matrix $\in \mathbb{R}^{L\times L}$\\
7:\quad\quad       $M_+,M_{\neq}$         $\gets$ Mask matrices $\in \mathbb{R}^{L\times L}$ for positive and non-equal pairs\\
9:\quad\quad       $L_{\theta_t}$         $\gets \sum (M_+\ast\big[S/\tau - \log(\sum M_{+}\ast e^{S/\tau_p} + \sum M_{\neq}\ast e^{S/\tau_n})\big])/\sum M_+$\\
10:\quad\quad\!\!\!$\theta_{t+1}$         $\gets \theta_t-\eta\nabla_{\theta_t}L_{\theta_t}$\\
11:\quad \textbf{end for} \\
12:\ \textbf{end for}
\end{minipage}
\end{algorithm}

\newpage

\subsection{Downstream Tasks}
\label{appendix:dwnstrm_settings}

\subsubsection{Task 1: Circuit Similarity Prediction}
\textbf{Setup.}
The model predicts the relative similarities between three circuits, including both analog and digital topologies.
The objective is to maximize prediction accuracy (\%) based on true similarity, which is determined by the number of shared labels.
A total of 6 labels include analog, digital, delay lines, amplifiers, logic gates, and oscillators.

Given a target circuit and two comparison circuits (total of three), the model outputs a three-dimensional probability distribution: whether the first circuit is more similar, the second circuit is more similar, or the two circuits are equally similar to the target circuit.
In each training step, one target circuit from the training dataset is sampled, and all permuted pairs across the $N$ number of circuits form a tensor of size $(N\times (N-1)=2950,\ \ 3\times graph\ feature\ dimension)$ for the training model.
Based on the number of shared labels, the logits indicating similarity comparisons are computed. Using these logits and the model output, the cross-entropy loss is minimized.

\textbf{Dataset.}
The training dataset consists of 50 circuits, including the 40 topologies from the pretraining dataset.
The test dataset contains 5 additional circuits, resulting in a total of 55 circuits.

\subsubsection{Task 2: Delay Prediction}
\textbf{Setup.}
The model predicts simulation results for five delay line circuits, which operate on digital signals.
Each simulation records two values: the time difference between the rising edges and the falling edges of the input and output clock signals.
The objective is to maximize the coefficient of determination ($R^2$) across all simulations, which measures how well the model’s predictions match the true simulation results.

\textbf{Dataset.}
For simulation, we used NGSPICE with the BSIM4 45nm technology, which serve as the open-source circuit simulator and technology file, respectively.
The delay line circuits used in Task 2 are not included in the pretraining dataset.
The simulation result dataset contains 45,000 device parameter combinations for each circuit topology.
We split the dataset into training, validation, and test sets with a ratio of 8:1:1.
The output dimension of the decoder model is 2.

\subsubsection{Task 3: OPAMP Metric Prediction}
\textbf{Setup.}
The model predicts simulation results for five operational amplifier (op-amp) circuits, which are analog circuits.
Each simulation records five metrics: power, DC voltage offset (V$_{\text{offset}}$), common-mode rejection ratio (CMRR), gain, and power supply rejection ratio (PSRR).
The objective is to maximize the coefficient of determination ($R^2$) across all simulations, which evaluates how well the model predictions align with the true simulation results.

\textbf{Dataset.}
NGSPICE with the BSIM4 45nm technology file is also used for Task 3.
The op-amp circuit topologies used in Task 3 are included in the pretraining dataset, but the simulation results are newly generated for this task.
The simulation dataset contains 60,000 device parameter combinations for each topology.
The circuit and testbench files are sourced from~\cite{li2024analoggym}.
The dataset is split into training, validation, and test sets using an 8:1:1 ratio.
The decoder network used in Task 3 is identical to that in Task 2, except the output dimension is set to 5.

\newpage

\subsection{Pretraining with DC voltage Prediction Task}
\label{appendix:pretrain_deepgen}
We pretrained DeepGen$_p$ using the DC voltage prediction task proposed in~\cite{hakhamaneshi2022pretraining, Hakhamaneshi2023_PretrainGNN}.
In the original work, the authors used two circuit topologies for pretraining: a resistor ladder and a simple operational amplifier (op-amp). They also trained separate GNNs for each topology.

However, to evaluate whether this supervised pretraining approach generalizes effectively, the pretraining process should include more than two circuit topologies simultaneously.
In our setting, we pretrained a single GNN on seven different circuit topologies: one resistor ladder, one current mirror, three simple RLC circuits, and two op-amps.
We used the same GNN architecture described in~\cite{hakhamaneshi2022pretraining, Hakhamaneshi2023_PretrainGNN} to ensure a fair comparison.

\subsection{Hyperparameters}
\label{appendix:hyperparameters}

\begin{table}[h]
\centering
\scriptsize
\begin{tabular}{cccccccc}
\hline\hline
\begin{tabular}[c]{@{}c@{}}Training\\ parameters\end{tabular} & \begin{tabular}[c]{@{}c@{}}DICE\\ Pretraining\end{tabular} & \begin{tabular}[c]{@{}c@{}}SimSiam\\ Pretraining\end{tabular} & \begin{tabular}[c]{@{}c@{}}NT-Xent\\ Pretraining\end{tabular} & \begin{tabular}[c]{@{}c@{}}DeepGen\\ Pretraining\end{tabular} & \begin{tabular}[c]{@{}c@{}}Downstream\\ Task 1\end{tabular} & \begin{tabular}[c]{@{}c@{}}Downstream\\ Task 2\end{tabular} & \begin{tabular}[c]{@{}c@{}}Downstream\\ Task 3\end{tabular} \\ \hline\hline
learning rate  & 0.0003 & 0.0003 & 0.0003  & 0.000005  & 0.00001  & 0.0001  & 0.0001   \\
batch size     & 1024 & 1024 & 1024    & 1024      & 50       & 2048    & 1024     \\
epochs         & 200 & 200 & 200     & 200       & 20000    & 300     & 300      \\
$(\tau_p,\tau, \tau_n)$  & (0.2, 0.05, 0.05) & (--,0.05,--) & -- & --  & --  & --  & -- \\
optimizer      & Adam & Adam & Adam & Adam & Adam & Adam & Adam      \\ \hline\hline
\end{tabular}
\end{table}

\begin{table}[h]
\centering
\scriptsize
\begin{tabular}{cccccc}
\hline\hline
Hyperparameters     & DICE & Encoder & Decoder & DeepGEN & ParaGraph  \\ \hline\hline
initial node feature dimension & 9 & -- & --   & 15      & 7          \\
initial edge feature dimension & 5 & -- & --   & 1       & 5          \\
hidden dimension    & 256  & 512     & 256     & 512     & 512        \\ 
dropout probability & 0.2  & --      & 0.3     & --      & --         \\ 
activation          & GELU & GELU    & GELU    & GELU    & GELU       \\ 
GNN type            & GIN  & GIN     & --      & GAT     & HGAT       \\ 
GNN depth           & 2    & depends & 0       & 3       & 3          \\ \hline\hline
\end{tabular}
\end{table}

\subsection{Compute Resources}
\label{appendix:compute_resources}
All experiments were conducted on a server running Ubuntu 20.04.6 LTS with Linux kernel 5.15. The machine is equipped with an Intel(R) Core(TM) i7-9700 CPU operating at 3.00 GHz, 62 GB of RAM, and a single NVIDIA Quadro RTX 6000 GPU with 24 GB of VRAM.

The DICE pretraining stage was executed on a single RTX A6000 GPU and required approximately 2 hours of wall-clock time per trial. Each downstream task was run on a single GPU as well, with Task 1 taking around 1 hour, Task 2 approximately 3 hours, and Task 3 about 8 hours per trial.

\newpage

\section{Limitations and Future Work}
\label{appendix:future}

Pretraining graph models that generalize across both analog and digital circuits remains in its early stages.
While we believe this work opens several promising research directions, it also has limitations that suggest avenues for future exploration.

First, more comprehensive comparisons could be made between different graph construction methods.
This is a challenging problem in its own right and should be analyzed with the same rigor as seen in recent work on molecules~\cite{wang2025a}.
Evaluations such as graph isomorphism tests may offer deeper insights into optimal circuit-to-graph mapping strategies.
Moreover, the definition of “devices” in this work could be broadened to include logic gates, allowing investigation into whether treating logic gates as fundamental units improves modeling efficiency.

Second, exploring alternative machine learning techniques could further enhance performance.
For instance, graph transformers may better capture global structural information.
In addition, other unsupervised graph representation learning methods—such as graph autoencoders or node/edge masking techniques—hold promise for future work.

Third, addressing more complex EDA tasks would provide stronger validation of the effectiveness of DICE.
While this work focuses on graph-level prediction tasks, analog circuit tasks like device parameter optimization~\cite{wang2020gcnrl, cao2022domain, gao2023rose, cao2024roseopt} pose greater challenges.
Similarly, digital circuit tasks such as PPA estimation, timing analysis, and congestion prediction involve significantly larger and more complex circuits than those used in this study.
Exploring these tasks would open valuable new directions for future research.

\section{Societal Impacts}
\label{appendix:societalimpacts}
Our work contributes to society by enabling more time- and cost-efficient circuit design. By reducing the need for labeled data and simulations, it can accelerate design workflows and make machine learning more accessible within the EDA community. Importantly, we view such automation as a collaborative tool that supports hardware designers, rather than a replacement for human expertise.

However, increased reliance on automation may lead to a decline in fundamental hardware understanding among future engineers and raise concerns about job displacement for traditional designers. Therefore, it is essential to ensure that efforts like ours are developed and deployed with the goal of empowering—rather than replacing—human designers.

\end{document}